\documentclass[11pt]{article}

\ifdefined\pdfobjcompresslevel
\fi

\usepackage{url}
\usepackage[utf8]{inputenc}
\usepackage{booktabs}
\usepackage{graphicx}
\usepackage[figuresright]{rotating}
\usepackage[section]{placeins}
\usepackage{textgreek}
\usepackage{multirow}
\usepackage{tabularx}
\usepackage{makecell}
\usepackage{float}
\usepackage{adjustbox}
\usepackage{siunitx}
\usepackage{geometry}
\usepackage{authblk}
\usepackage{lmodern}
\usepackage{bm}
\usepackage{helvet}
\usepackage{amsmath}
\usepackage{hyperref}
\usepackage{microtype}
\usepackage{svg}
\usepackage{multirow}
\usepackage{wrapfig}
\usepackage{xspace}
\usepackage[capitalise,noabbrev,nameinlink]{cleveref}
\crefformat{equation}{#2Equation~#1#3}
\usepackage{pgfplots,pgfplotstable}
\usepackage{subcaption}
\usepackage{color}
\usepgfplotslibrary{colorbrewer,groupplots}
\usetikzlibrary{positioning}
\pgfplotsset{compat=1.18}
\definecolor{visible-blue}{rgb}{0.286, 0.525, 0.910}
\definecolor{tabfirst}{rgb}{1, 0.7, 0.7} % red
\definecolor{tabsecond}{rgb}{1, 0.85, 0.7} % orange
\definecolor{tabthird}{rgb}{1, 1, 0.7} % yellow

\newcommand{\OOM}[1]{-}
\usepackage{tocloft}
\begin{document}
\thispagestyle{empty}

\title{\Large \bf Nimbus: A Unified Embodied Synthetic Data Generation Framework}

\author{Zeyu He$^1$, Yuchang Zhang$^1$, Yuanzhen Zhou$^1$, Miao Tao$^1$, Hengjie Li$^{1,2}$\thanks{Corresponding author.}, Hui Wang$^1$, Yang Tian$^1$, \mbox{Jia~Zeng$^1$}, Tai Wang$^1$, Wenzhe Cai$^1$, Yilun Chen$^1$, Ning Gao$^1$, Jiangmiao Pang$^1$}

\affil{$^1$Shanghai Artificial Intelligence Laboratory}
\affil{$^2$Shanghai Innovation Institute}

\date{}

\maketitle

\begin{abstract}
Scaling data volume and diversity is critical for generalizing embodied intelligence. 
While synthetic data generation offers a scalable alternative to expensive physical data acquisition, existing pipelines remain fragmented and task-specific. 
This isolation leads to significant engineering inefficiency and system instability, failing to support the sustained, high-throughput data generation required for foundation model training. 
To address these challenges, we present Nimbus, a unified synthetic data generation framework designed to integrate heterogeneous navigation and manipulation pipelines. 
Nimbus introduces a modular four-layer architecture featuring a decoupled execution model that separates trajectory planning, rendering, and storage into asynchronous stages. 
By implementing dynamic pipeline scheduling, global load balancing, distributed fault tolerance, and backend-specific rendering optimizations, the system maximizes resource utilization across CPU, GPU, and I/O resources. 
Our evaluation demonstrates that Nimbus achieves a $2$--$3\times$ improvement in end-to-end throughput compared to unoptimized baselines and ensuring robust, long-term operation in large-scale distributed environments. 
This framework serves as the production backbone for the InternData suite, enabling seamless cross-domain data synthesis.
\end{abstract}

\clearpage
\tableofcontents
\clearpage

\section{Introduction}

Progress in embodied intelligence depends fundamentally on two core capabilities: navigation and manipulation. 
These primitives enable agents to move autonomously through environments and interact physically with objects,
establishing them as the fundamental building blocks of Artificial General Intelligence (AGI). 
Existing works ($\pi_{0.5}$~\cite{pi05}, Gen-0 \cite{generalist2025gen0}) 
have demonstrated that scaling data scale and diversity further could accelerate the capability growth of models. 
However, most of these efforts rely on real-world data collection, where data acquisition remains prohibitively expensive. 
Specifically, the high capital cost of hardware deployment and the temporal overhead of collector training prevent physical pipelines from satisfying the high-capacity, multimodal demands of modern foundation model training.

Synthetic data pipelines (\cite{navdp}, \cite{gao2025genmanip}, \cite{interndata2025a1}) offer a scalable alternative to bypass the physical data bottlenecks 
by generating controllable, diverse data in virtual environments.
However, existing synthetic data pipelines lack a unified framework, which restricts current solutions to being tailored exclusively to either navigation or manipulation tasks.
This deficiency gives rise to two core issues: first, inefficiency, marked by engineering redundancy and the lack of generalizability in optimization and acceleration strategies; 
second, instability, where the absence of standardized cluster scheduling and fault tolerance mechanisms hinders the sustained, long-cycle data generation required for large-scale datasets.

To bridge this gap, we present Nimbus, a unified synthetic data generation framework designed to integrate heterogeneous navigation and manipulation pipelines.
Concretely, Nimbus is built around a layered design and a set of multi-layer optimizations.
At the framework level, we introduce a four-layer architecture that cleanly separates scheduling optimization and fault tolerance (\textit{Schedule Opt Layer}), a unified \textit{Load-Plan-Render-Store} lifecycle runner abstraction (\textit{Stage Runner Layer}), interface-driven reusable components for navigation and manipulation (\textit{Components Layer}), and a high-performance runtime that integrates an environment based on Ray~\cite{moritz2018ray} and different simulator and renderer backends (\textit{Backend Opt Layer}).
This modularization enables the same scheduling and optimization primitives to be applied across heterogeneous pipelines (e.g., InternData-N1 navigation and InternData-A1/M1 manipulation) without rewriting scenario logic.

At the Schedule Opt Layer, we implement a two-level optimization stack consisting of a pipeline parallelism and a distributed optimization.
In the pipeline parallelism, we decouple the traditional monolithic pipeline into an asynchronous execution model.
Trajectory planning (CPU-bound), rendering (GPU-bound), and storage (I/O-bound) are executed in independent worker pools with dynamic pipeline scheduling.
This design mitigates blocking and maximizes CPU, GPU, and I/O utilization.
In the distributed optimization, we target cluster-scale high efficiency and availability.
We employ a global balancer and per-worker supervisors to enforce load balancing, liveness monitoring, and automatic recovery.
Finally, at the Backend Opt Layer, we optimize the critical rendering paths for major backends (Gaussian Splatting~\cite{kerbl3Dgaussians}, Blender, and Isaac Sim).
Techniques such as accelerated rasterization and batched/stacked rendering are applied to further increase per-process throughput.
Together, these designs deliver a $2$--$3\times$ end-to-end throughput improvement over an unoptimized baseline and enable stable, sustained data generation in distributed environments.

In summary, the primary contributions of our work are:
\begin{itemize}
    \item We propose Nimbus, a unified framework that integrates the fragmented pipelines of navigation and manipulation. 
    This design enables seamless cross-domain data synthesis, serving as the production backbone for the InternData suite \cite{interndata2025a1, internvla-n1, internvla-m1}.
    \item We decouple pipeline stages and implement dynamic pipeline scheduling with rendering optimization to maximize resource utilization. 
    Compared to the baseline without optimization, this design boosts end-to-end throughput by 2--3$\times$. 
    \item We also introduce a high-availability design that supports large-scale cluster deployment.
    It incorporates robust fault recovery mechanisms to ensure stability over sustained operation, effectively resolving the reliability bottlenecks inherent in traditional data pipelines.
\end{itemize}

The remainder of this paper is organized as follows. 
Section~\ref{sec:background} details the InternData series datasets and their generation pipeline. 
Section~\ref{sec:relatedwork} analyzes existing literature across navigation, manipulation, and related frameworks. 
Section~\ref{sec:architecture} describes the Nimbus architecture and its integration strategy. 
Section~\ref{sec:optimizations} presents our optimization techniques, including stage decoupling, scheduler design, and distributed fault tolerance. 
Section~\ref{sec:experiment} evaluates the framework's performance, and Section~\ref{sec:conclusion} concludes with future directions.

\section{Background}
\label{sec:background}

\subsection{InternData-N1}
InternData-N1~\cite{internvla-n1,streamvln,navdp} is a large-scale synthetic dataset for Vision-Language Navigation (VLN), covering over 3,000 indoor scenes with 53.5 million first-person view (FPV) images and 800,000 language instructions (approximately 4,840 km of trajectories). It includes three complementary subsets (VLN-N1/VLN-CE/VLN-PE) to support general navigation pre-training, fine-grained instruction following, and sim-to-real transfer, respectively.

\begin{figure}[t]
    \centering
    \includegraphics[width=\linewidth]{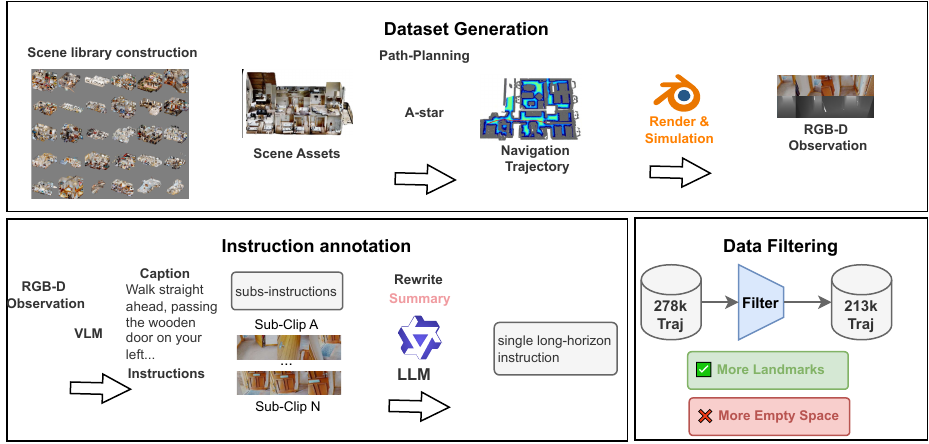}
    \caption{InternData-N1 pipeline}
    \label{fig:interndata-n1-pipeline}
\end{figure}

Figure~\ref{fig:interndata-n1-pipeline} illustrates the synthesis workflow, which can be summarized as:
\begin{itemize}
    \item \textbf{Scene library construction.} Aggregate six open-source indoor scene repositories (Replica, Matterport3D, Gibson, 3D-Front, HSSD, and HM3D) to cover diverse room layouts and scene categories.
    \item \textbf{Path planning.} Generate collision-free, smooth trajectories via a three-stage procedure: (i) construct a Euclidean Signed Distance Field (ESDF) per floor and use A-star to plan an initial global path between randomly sampled start--goal pairs; (ii) refine turning points based on ESDF to maintain obstacle clearance; and (iii) apply B\'ezier smoothing to ensure continuous motion.
    \item \textbf{Observation rendering.} Render the planned trajectories frame-by-frame with BlenderProc to obtain FPV RGB images and depth maps.
    \item \textbf{Instruction annotation.} Detect keyframes from geometric cues (e.g., sharp turns or passing landmarks) and split trajectories into sub-segments; use a multimodal model (LLaVA-OneVision) to generate fine-grained step instructions; then rewrite and summarize the sub-instructions with a language model (Qwen3-72B) into a single long-horizon instruction.
    \item \textbf{Data filtering.} Remove low-quality samples using landmark density and scene information metrics.
\end{itemize}

\subsection{InternData-A1}
InternData-A1~\cite{interndata2025a1} is a high-fidelity synthetic manipulation dataset for pre-training generalist Vision-Language-Action (VLA) policies~\cite{internvla-a1}. It spans four robot morphologies (Genie-1, Franka Panda, AgileX Split Aloha, and ARX Lift-2) and covers 18 skills across 227 indoor scenes, totaling 7,433 hours of interaction data.

\begin{figure}[t]
    \centering
    \includegraphics[width=\linewidth]{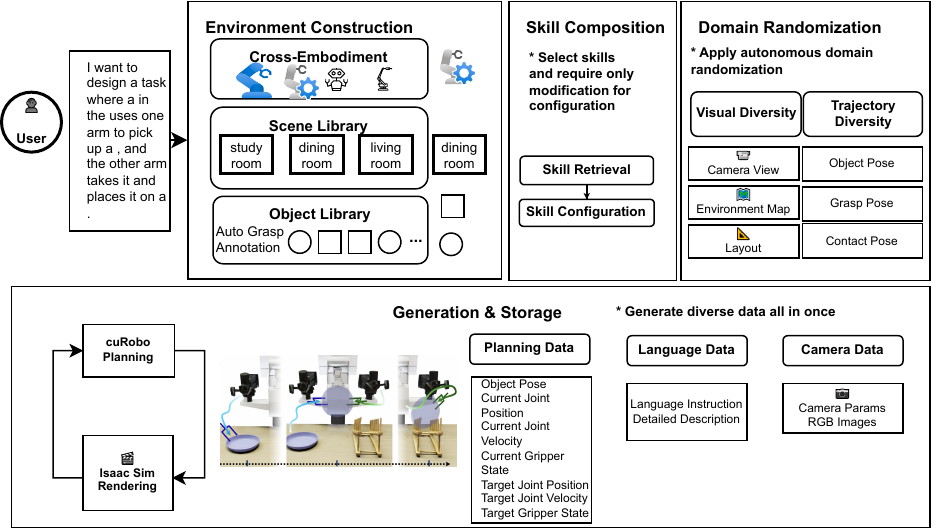}
    \caption{InternData-A1 pipeline}
    \label{fig:interndata-a1-pipeline}
\end{figure}

As shown in Figure~\ref{fig:interndata-a1-pipeline}, InternData-A1 is synthesized through a fully decoupled and composable pipeline:
\begin{itemize}
    \item \textbf{Environment construction.} Given a task template, retrieve task-relevant robots, scenes, and objects from an asset library: robots are provided as validated USD embodied models; scenes are sourced from GRUtopia GRScenes-100 with annotated manipulation areas; objects include rigid, articulated, deformable, and fluid categories with high-fidelity physical models (e.g., AnyGrasp-generated grasp poses for rigid objects and dedicated simulators for deformables/fluids).
    \item \textbf{Skill composition.} Build tasks by selecting and composing atomic skills from a skill library. Each skill is implemented as a scripted policy that maps object/robot states and constraints to a sequence of end-effector 6D waypoints, decoupling high-level logic from low-level interpolation and control (e.g., Pick, Place, Push, Goto-pose, and Gripper-action).
    \item \textbf{Domain randomization.} Apply systematic randomization on both appearance and dynamics to narrow the sim-to-real gap: perturb camera extrinsics (main and wrist cameras), sample diverse HDR environment maps and lighting parameters, swap instances within object categories, and randomize table/background layouts; additionally randomize initial robot/object poses and stochasticity in grasp/contact regions (e.g., sampling from top-confidence AnyGrasp candidates).
    \item \textbf{Trajectory generation and storage.} Given waypoint sequences, use CuRobo for joint-space planning and dense interpolation; record multi-view RGB observations with camera parameters, robot states and control commands, object metadata, and language instructions (optionally exporting depth, grounding labels, and boxes). Only successful rollouts are rendered and written to the dataset, which is finally converted into the LeRobot format.
\end{itemize}

\subsection{InternData-M1}
InternData-M1~\cite{internvla-m1, gao2025genmanip} is a synthetic tabletop manipulation dataset targeting long-horizon reasoning and spatial grounding. It contains approximately 244,000 simulation demonstrations with dense action and geometric annotations, and incorporates over 80,000 open-vocabulary objects to promote generalization.

\begin{figure}[t]
    \centering
    \includegraphics[width=\linewidth]{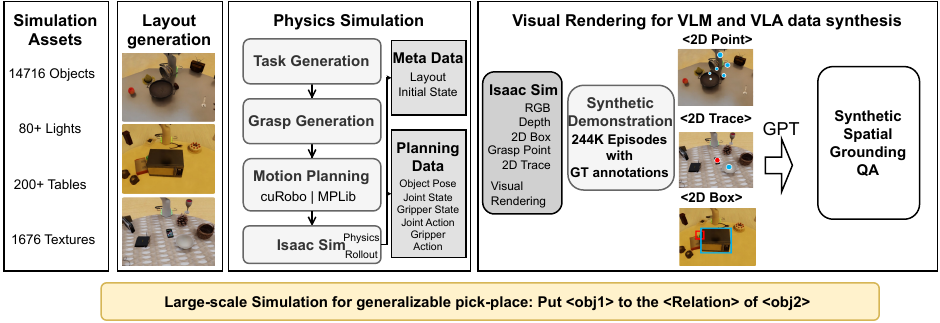}
    \caption{InternData-M1 pipeline}
    \label{fig:interndata-m1-pipeline}
\end{figure}

Figure~\ref{fig:interndata-m1-pipeline} summarizes the LLM-driven simulation workflow, which proceeds as follows:
\begin{itemize}
    \item \textbf{Simulation assets.} Construct a large asset library in Isaac Sim, including over 14,000 labeled objects, 200 tables, and nearly 1,700 texture/lighting configurations to provide broad physical and visual diversity.
    \item \textbf{Physics simulation and task synthesis.} Synthesize trajectories by randomizing object layouts and illumination, and by using privileged signals (e.g., meshes and robot states) to compute grasp candidates and motion plans. Each trajectory is validated in closed-loop execution and checked by a scene-graph validator; only successful and physically consistent rollouts are retained.
    \item \textbf{Decoupled planning and visual rendering.} Decouple planning from rendering by first recording structured planning traces (e.g., joint states and object poses) and then replaying them under randomized viewpoints, materials, and lighting. Camera calibration is performed via ArUco markers to align intrinsic/extrinsic parameters with real sensors, while the renderer produces RGB images and dense spatial supervision (e.g., 2D boxes, end-effector traces, and precise keypoints).
    \item \textbf{VLM/VLA data packaging.} Convert intermediate representations into a unified visual question answering (VQA) format for VLM pre-training. By pairing actions with natural language instructions and spatial annotations, the pipeline derives auxiliary supervision for affordance recognition, trajectory prediction, and multi-step planning, bridging semantic reasoning and embodied execution.
\end{itemize}

\section{Related Work}
\label{sec:relatedwork}
\subsection{Navigation Data Generation}
Embodied navigation tasks predominantly rely on pre-constructed 3D environments and static task distributions. While mainstream data sources leverage large-scale indoor scans such as Matterport3D~\cite{chang2017matterport3d} and Habitat-Matterport 3D (HM3D)~\cite{hm3d}, their acquisition orchestrates a resource-intensive pipeline involving specialized hardware, physical staging, and complex post-processing. 

In the VLN domain, the dominant approach involves pairing static environmental assets with human-generated annotation. Benchmarks such as R2R~\cite{anderson2018vision} and RxR~\cite{ku2020room} are constrained to limited environments, relying on labor-intensive crowdsourcing pipelines to generate instructions for pre-defined paths. Although successors like CVDN~\cite{thomason2019vision} and REVERIE~\cite{qi2020reverie} incorporate multi-turn dialogues and referring expressions to enhance semantic complexity, they incur significant overhead due to manual intervention in scene selection, trajectory sampling, and verification. Consequently, the marginal cost of scaling these datasets remains prohibitively high.

Conversely, ObjectNav~\cite{objectnav} and related goal-driven tasks demand rigorous scene semantics, necessitating precise object lists, instance segmentation, and category labels to enable offline task sampling. While platforms like Habitat automate this process by applying rule-based generation to scanned assets (e.g., HM3D), the underlying logic is often tightly coupled to specific codebases. The absence of unified abstractions for task templates, trajectory sampling, and metadata organization prevents modularity, significantly hindering cross-project reusability.

Ultimately, contemporary navigation data generation follows a hybrid real-world scanning and simulation workflow. Construction costs are front-loaded in physical scanning and post-processing, while task generation remains bottlenecked by manual annotation or brittle, ad-hoc scripts. As a result, dataset updates manifest as discrete, static releases rather than a continuous, evolutionary production process.

\subsection{Manipulation Data Generation}
In contrast to navigation tasks, data generation for manipulation has evolved along three distinct vectors over the past three years: (1) large-scale physical teleoperation, representing the traditional real-world demonstration paradigm; (2) Universal Manipulation Interface (UMI)-style approaches (e.g., UMI, Fast-UMI)~\cite{chi2024universal, zhaxizhuoma2025fastumi} that decouple human operators from robot morphology to capture robot-like trajectories in the wild; and (3) synthetic data generation, which orchestrates simulation environments via demonstration augmentation or rule-based batch processing. 

\paragraph{Physical Teleoperation and Multi-Robot Aggregation.}
Cross-institutional initiatives, such as Open X-Embodiment~\cite{padalkar2023open}, aggregate data from 34 laboratories and over 60 sub-datasets to construct massive repositories of real-world trajectories. Concurrently, systems like $\pi_0$~\cite{black2024pi0} and RoboMIND~\cite{wu2024robomind} enforce unified collection protocols within single institutions to produce standardized, multi-robot corpora. Recent efforts, including AgiBotWorld~\cite{bu2025agibot} and Galaxea~\cite{jiang2025galaxea}, have expanded this paradigm to diverse open-world scenarios yielding high-resolution, long-horizon trajectories. While these works significantly scale real-world data, they face systemic scalability bottlenecks: they necessitate the dedicated occupation of hardware resources, rely heavily on expert teleoperators, and incur substantial overhead for compliance, privacy, and heterogeneous platform integration.

To mitigate engineering friction, immersive interfaces like IRIS~\cite{jiang2025iris} leverage XR and point cloud rendering to unify cross-platform teleoperation views. Meanwhile, industrial players (e.g., Tesla, Figure AI) orchestrate large-scale fleets to collect proprietary behavior data. However, these industrial pipelines remain closed-source, obscuring their internal orchestration logic and preventing external researchers from leveraging their system abstractions.

\paragraph{UMI-like Interfaces and Proxy Devices.}
A second method decouples the operator from the robot entirely. Approaches such as UMI, DexUMI~\cite{xudexumi}, and DexCap~\cite{wang2024dexcap} employ proxy devices to capture in-the-wild demonstrations that are offline-mapped to target robot morphologies. UMI, for instance, utilizes a low-cost handheld interface with multimodal sensors, enabling non-experts to demonstrate complex dynamic manipulations. Fast-UMI further abstracts the hardware dependency, employing commercial tracking to simplify calibration and utilizing a specialized toolchain for data conversion.

While UMI-like interfaces preserve real-world physics without monopolizing expensive robot hardware, they remain linearly bounded by human demonstration time. Furthermore, the translation from proxy device to robot imposes significant engineering complexity, necessitating rigorous coordinate mapping, spatiotemporal alignment, and unified control interfaces.

\paragraph{Synthetic Data: Augmentation and Rule-Driven Generation.}
Synthetic generation approaches trade physical realism for scalability, broadly categorized into demonstration augmentation and rule-driven generation. 
MimicGen~\cite{mandlekar2023mimicgen} exemplifies the augmentation approach, procedurally expanding a small seed set of human demonstrations into thousands of trajectories via sub-task segmentation and scene perturbation. RoboCasa~\cite{nasiriany2024robocasa} and its successor RoboCasa365 integrate this logic into a high-fidelity home simulation platform, effectively operating as a demonstration factory.
Conversely, rule-driven data generation like InternData-M1 pipeline and InternData-A1 pipeline to generate massive datasets from pre-defined task scripts and physical constraints. 
Unlike the fixed benchmarks of RLBench~\cite{james2020rlbench} or ManiSkill~\cite{gu2023maniskill2}, these works emphasize synthetic data as a service, exposing interfaces that allow researchers to reuse asset construction logic.

Simulation synthesis converts the marginal cost of data generation into computational overhead. While this enables the production of millions of trajectories with negligible human intervention, it necessitates robust verification mechanisms to filter physically implausible data and bridge the sim-to-real gap.

\paragraph{Summary: The Dataset-Centric Bottleneck.}
The three prevailing paradigms reveal a critical systemic gap. Physical and UMI-based routes are constrained by hardware and human labor, while simulation shifts the bottleneck to asset and script construction. Crucially, the current landscape is dataset-centric: once a corpus is finalized, the underlying generation pipeline is effectively frozen. Researchers are restricted to consuming static datasets, unable to incrementally extend task definitions or sensor configurations without re-engineering the entire workflow. This lack of a reusable, orchestratable system architecture motivates the design of Nimbus.

\subsection{System Frameworks for Data Generation}
From a systems perspective, existing frameworks fall into three categories: task-specific generation tools, simulation platforms, and general-purpose workflow engines.

The first category, represented by MimicGen, formalizes data expansion into a programmable pipeline but restricts control to task-level rule modifications.

The second category, including RoboCasa, integrates assets and collection tools into unified platforms. While effective for their specific domains, these systems enforce rigid boundaries; their internal pipelines are tightly coupled to specific simulation backends, hindering migration to new modalities or tasks.

The third category comprises general-purpose orchestrators like Apache Airflow~\cite{airflow} and Prefect~\cite{prefect}. While mature for industrial ETL, these systems treat tasks as black-box operators and lack native abstractions for the robot-environment-sensor topology. They are optimized for batch processing rather than the real-time, stateful control loops required for embodied AI. Furthermore, they track file-level metadata rather than the semantic linkage between trajectories, scene states, and policy versions.

In distributed deep learning, pipeline-parallel training schedules such as Megatron-LM’s 1F1B~\cite{megatron} and Zero-Bubble Pipeline Parallelism~\cite{qi2024zero} improve throughput by optimizing the execution order of micro-batches across layer-partitioned pipeline stages under synchronous training semantics, aiming to reduce pipeline bubbles.

These methods are formulated around a fixed set of pipeline stages with explicit inter-stage activation/gradient dependencies and consistency constraints required for correct gradient computation, and their bubble-related optimality is typically analyzed under a stage-time cost model, while real deployments may suffer from execution-time variance and stragglers.

In contrast, data-collection sessions involves asynchronous, event-driven orchestration with dynamic participation, failures/retries, and heterogeneous latencies, which calls for a different scheduling abstraction and objective than layer/microbatch-level training pipelines.

Consequently, a gap remains for a general-purpose framework that spans real and simulated environments, unifies collection and generation, and provides native orchestration for embodied tasks. Our Nimbus addresses this by introducing system-level abstractions specifically designed for the lifecycle of embodied AI data.

\section{Architecture}
\label{sec:architecture}

\begin{figure}[t]
    \centering
    \includegraphics{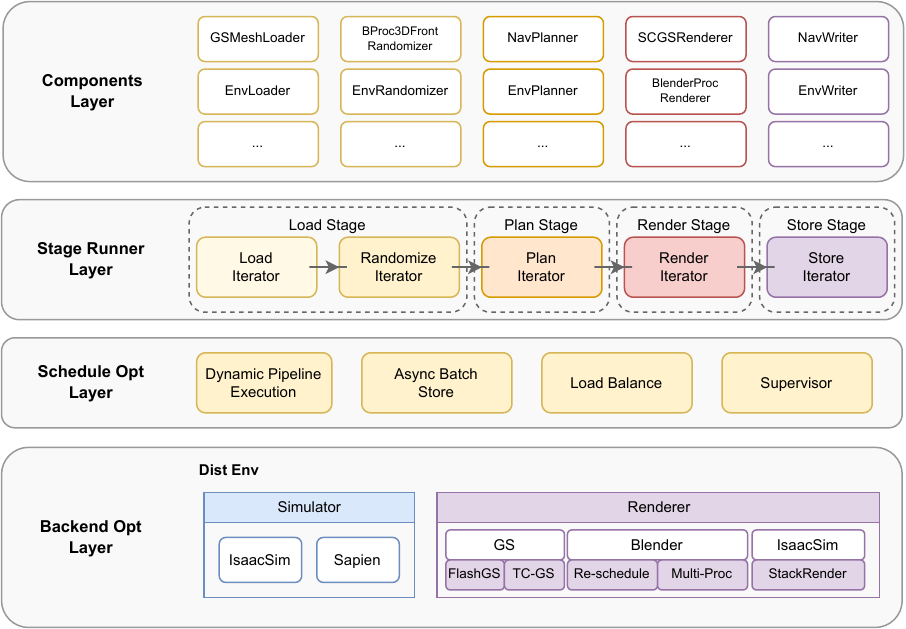}
    \caption{Nimbus Layered Architecture. The framework separates domain logic (Components Layer), workflow abstraction (Stage Runner Layer), control logic (Schedule Opt Layer), and execution runtime (Backend Opt Layer).}
    \label{fig:arch}
\end{figure}

Nimbus adopts a modular, four-layer architecture designed to unify heterogeneous synthetic data pipelines while maximizing resource efficiency. 
As illustrated in Figure~\ref{fig:arch}, the framework is organized into the Components Layer, Stage Runner Layer, Schedule Opt Layer, and Backend Opt Layer. 
This layered design decouples high-level orchestration from low-level execution, enabling unified scheduling and optimization primitives across diverse navigation and manipulation tasks.

\subsection{Stage Runner Layer}
The Stage Runner Layer defines a standardized \textit{Load-Plan-Render-Store} lifecycle, abstracting the generation workflow into four stages. 
The \textbf{Load Stage} handles asset ingestion and domain randomization to expand data diversity. 
The \textbf{Plan Stage} generates physically valid motion sequences or action plans based on task specifications. 
The \textbf{Render Stage} visualizes these sequences into high-fidelity multimodal sensor observations (RGB, Depth, etc.). 
Finally, the \textbf{Store Stage} manages data serialization, aggregating heterogeneous outputs into a unified persistence format.

\begin{table}[t]
\centering
\caption{Interface Specifications of Core Abstract Base Classes}
\label{tab:components}
\resizebox{\textwidth}{!}{
\begin{tabular}{l l l l p{5cm}}
\toprule
\textbf{Abstract Class} & \textbf{Core Interface} & \textbf{Input Format} & \textbf{Output Format} & \textbf{Responsibilities} \\
\midrule
\texttt{BaseLoader} & \texttt{load\_asset(config)} & \texttt{Config} & \texttt{Scene} & Asset ingestion and integrity verification. \\ 
\midrule
\texttt{BaseRandomizer} & \texttt{randomize(scene)} & \texttt{Scene} & \texttt{Scene} & Domain randomization. \\ 
\midrule
\texttt{BasePlanner} & \texttt{gen\_sequence(scene)} & \texttt{Scene} & \texttt{Sequence} & Generation of feasible trajectories or action plans. \\ 
\midrule
\texttt{BaseRenderer} & \texttt{gen\_obs(seq)} & \texttt{Sequence} & \texttt{Observation} & Multimodal synthesis (RGB, Depth, Segmentation, etc.). \\ 
\midrule
\texttt{BaseWriter} & \texttt{save\_data(seq, obs)} & \texttt{Sequence, Observation} & \texttt{File (Parquet)} & Data serialization and unified persistence. \\
\bottomrule
\end{tabular}
}
\end{table}

To standardize execution within each stage, we define five abstract base iterators that enforce strict input/output protocols, as detailed in Table~\ref{tab:components}. 
In the Load Stage, we provide \texttt{BaseLoader} and \texttt{BaseRandomizer} to collectively construct scenes. 
Specifically, \texttt{BaseLoader} ingests configuration to output a scene object, which is then processed by \texttt{BaseRandomizer} to modify attributes for domain randomization. 
For the remaining stages, \texttt{BasePlanner} takes the scenes to generate trajectory sequences; 
\texttt{BaseRenderer} synthesizes observations from these sequences; 
and \texttt{BaseWriter} serializes sequences and observations to storage. 
By chaining these iterators, Nimbus orchestrates the end-to-end synthetic data generation flow in a unified manner.

\subsection{Components Layer}
\label{sec:components}

\begin{figure}[t]
    \centering
    \includegraphics{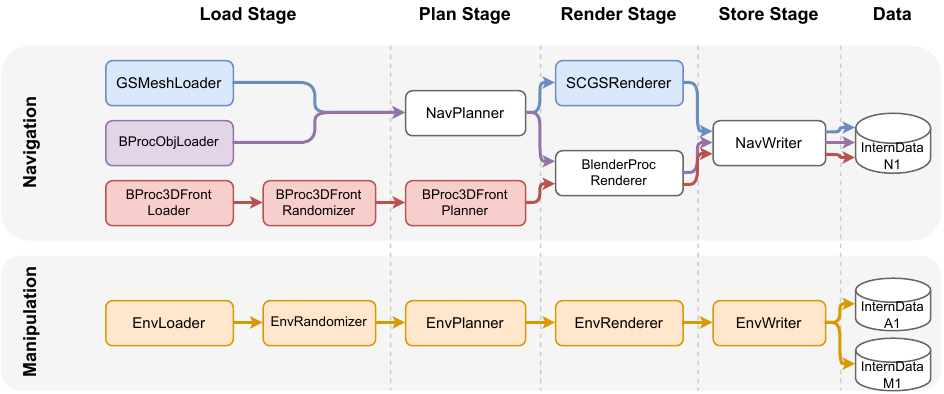}
    \caption{Example of Nimbus implementing data generation for different data pipelines via component combination}
    \label{fig:components_pipeline}
\end{figure}

The Components Layer implements specific logic for navigation and manipulation within the unified interfaces defined by the Stage Runner. 
This design addresses the fragmentation of existing pipelines by enforcing strict interface contracts, enabling code reuse and seamless integration. 
As shown in Figure~\ref{fig:components_pipeline}, Nimbus integrates distinct Navigation and Manipulation workflows through flexible component composition.

\subsubsection{Navigation Components \& Integration}
The Navigation components adapt the generic lifecycle to the InternData-N1 pipeline, addressing the heterogeneity between mesh-based and Gaussian Splatting assets.

\paragraph{Versatile Component Suite.}
We implement a comprehensive suite of components to handle diverse data formats across the generation lifecycle.
In the Load Stage, we provide \texttt{GSMeshLoader} for proxy meshes aligned with GS assets, \texttt{BProcObjLoader} for standard .obj files, and \texttt{BProc3DFrontLoader} for parsing semantic layouts from the 3D-FRONT dataset.
To expand data diversity, \texttt{BProc3DFrontRandomizer} applies domain-specific augmentations, including lighting, texture, and pose randomization.
In the Plan Stage, \texttt{NavPlanner} generates collision-free trajectories using A-star pathfinding. 
This could be extended by dataset-specific logic (e.g., \texttt{BProc3DFrontPlanner}) to enforce layout constraints in structured environments.
For the Render Stage, we employ \texttt{SCGSRenderer} for high-fidelity GS rasterization and \texttt{BlenderProcRenderer} for photorealistic mesh rendering.
Finally, in the Store Stage, \texttt{NavWriter} aggregates trajectory poses, multimodal observations, and scene metadata, serializing them into the unified InternData-N1 format.

\paragraph{Pipeline Integration Strategy.}
To reconcile the distinct characteristics of GS and Mesh assets, we employ differentiated component composition strategies, as illustrated in Figure~\ref{fig:components_pipeline}.
For Mesh assets, the pipeline could be configured based on format complexity.
For GS assets, which offer superior rendering fidelity but lack the geometric information required for path planning, we implement a proxy-based pipeline: \texttt{GSMeshLoader} $\rightarrow$ \texttt{NavPlanner} $\rightarrow$ \texttt{SCGSRenderer} $\rightarrow$ \texttt{NavWriter}.
Specifically, \texttt{GSMeshLoader} ingests a proxy mesh aligned with the GS coordinate system, enabling \texttt{NavPlanner} to compute physically valid trajectories.
These trajectories are subsequently rendered using \texttt{SCGSRenderer}, effectively combining geometric validity with visual fidelity. 

\subsubsection{Manipulation Components \& Integration}
The Manipulation components unify the heterogeneous workflows of InternData-A1 pipeline and InternData-M1 pipeline.
In contrast to the flexible composition used for navigation, we employ \textit{Unified Adapter Components} with a standardized \textit{Workflow API} to manage the complexity of robotic simulation.

\paragraph{Unified Adapter Components.}
We define a suite of \textit{Env} components (e.g., \texttt{EnvLoader}, \texttt{EnvPlanner}, etc.) that serve as abstraction wrappers around underlying simulators such as IsaacSim and Sapien.
These components map the domain-specific operations of InternData-A1 and InternData-M1 pipelines to the standardized Nimbus lifecycle stages.

\paragraph{Workflow API.}
The implementation of Unified Adapter Components is underpinned by a standardized Workflow API, which provides three key capabilities:
\begin{itemize}
    \item \textbf{Encapsulation:} It encapsulates pipeline logic and simulation interfaces, shielding the component layer from backend discrepancies.
    \item \textbf{Decoupling:} By defining core interfaces such as \texttt{reset}, \texttt{randomization}, \texttt{generate\_seq}, \texttt{seq\_replay}, and \texttt{save}, it strictly separates detailed generation logic from the framework's scheduling and optimization mechanisms.
    \item \textbf{Extensibility:} Developers implement specific workflows, such as \texttt{GenManipWorkflow} for InternData-A1 pipeline and \texttt{SimBoxWorkflow} for InternData-M1 pipeline, by adhering to this interfaces. 
    The framework's Env components invoke these standardized methods, remaining agnostic to the underlying implementation details.
\end{itemize}

\subsection{Schedule Opt Layer}
The Schedule Opt Layer serves as the control plane, responsible for global resource management and fault tolerance. 
It employs four key optimizations to address monolithic execution inefficiencies and cluster instability:

\begin{itemize}
    \item \textbf{Dynamic Pipeline Execution:} Leveraging the decoupling of trajectory planning and rendering, we implement a dynamic pipelining mechanism. This design eliminates inter-stage blocking, effectively masking computation latency and maximizing throughput.
    \item \textbf{Asynchronous Batch Storage:} To mitigate I/O overhead, we offload data persistence to asynchronous threads. 
    By batching write operations, this module isolates storage latency from the critical computation path.
    \item \textbf{Load Balancing:} Addressing resource skew in distributed environments, the load balancer dynamically schedules tasks to saturate cluster capacity. 
    This strategy effectively mitigates the straggler effect and ensures uniform resource utilization.
    \item \textbf{Supervisor:} Providing fine-grained fault tolerance, the supervisor monitors task liveness in real-time. 
    It automatically detects failures and triggers recovery routines to guarantee continuous system availability.
\end{itemize}

Detailed implementation of these optimization is presented in Section~\ref{sec:optimizations}.

\subsection{Backend Opt Layer}
The Backend Opt Layer provides the high-performance runtime environment. 
It integrates a Ray-based environment with optimized renderer backends. 
Key optimizations include accelerated rasterization for Gaussian Splatting, RT-cores and tensor-cores optimization for Blender, and batched rendering for Isaac Sim. 
These backend-specific enhancements work in concert with the upper layers to saturate hardware capabilities.
\section{Design of Multi-Layer Optimization}
\label{sec:optimizations}

Monolithic synthetic data pipelines often couple trajectory planning and visual rendering into a synchronous execution unit. 
While convenient for prototyping, this coupling introduces severe inefficiencies at scale. 
First, the tight coupling of planning and rendering leads to computation waste: 
invalid trajectories generated during the planning phase still trigger rendering, consuming resources unnecessarily.
Second, the resource profiles of the two stages diverge significantly: planning is CPU-bound while rendering is GPU-bound.
Serial execution forces one resource type to idle while the other is active.
It results in severe hardware underutilization. 
Moreover, as deployment scales, partial failures become inevitable, mandating robust fault tolerance to ensure continuous availability.

\begin{figure}[t]
    \centering
    \includegraphics[width=\linewidth]{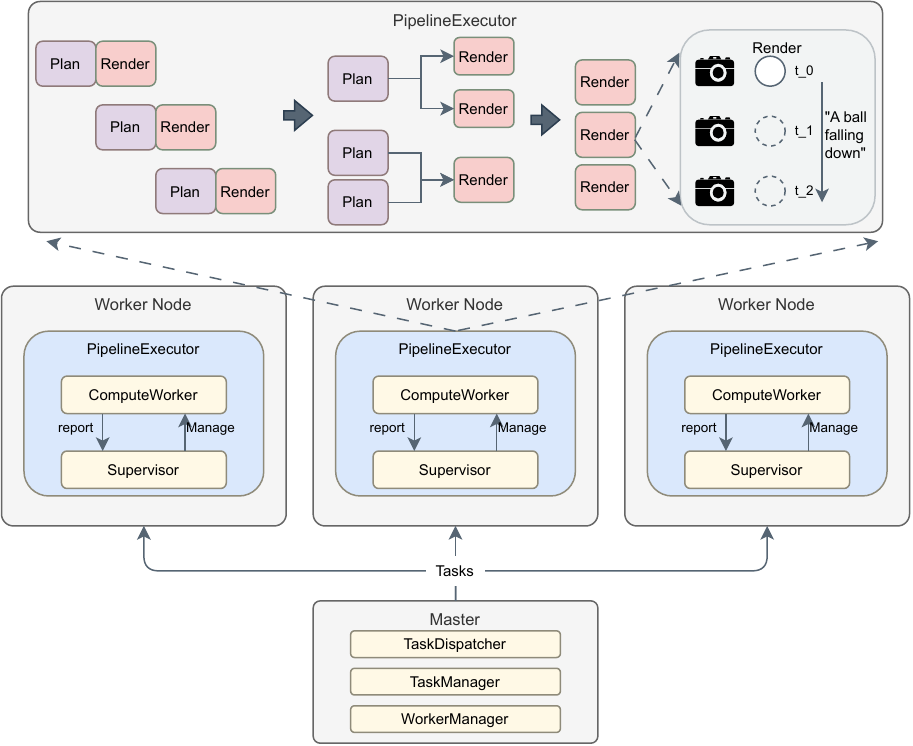}
    \caption{Schematic diagram of multi-layer optimization of the Nimbus framework}
    \label{fig:multi_layer_optimization}
\end{figure}

To mitigate these bottlenecks, Nimbus implements a multi-layer optimization strategy mentioned in the Schedule Opt Layer and Backend Opt Layer defined in Section~\ref{sec:architecture}.
Figure~\ref{fig:multi_layer_optimization} illustrates the multi-layer optimization strategy.
The top part depicts the \textit{Pipeline Parallelism} (Section~\ref{sec:pipeline_parallel}) and \textit{Renderer Optimization} (Section~\ref{sec:backend_opt}), which decouple the serial execution into an asynchronous model and accelerate rendering to maximize aggregate throughput.
The bottom part demonstrates the \textit{Distributed Optimization} (Section~\ref{sec:distributed_opt}), employing global load balancing with supervisors to ensure cluster-scale efficiency and availability.

\subsection{Schedule Opt Layer: Pipeline Parallelism}
\label{sec:pipeline_parallel}
By decoupling the generation lifecycle into asynchronous stages, we implement Dynamic Pipeline Execution and Asynchronous Batch Storage. 
This design employs fine-grained \texttt{ComputeWorker} encapsulation and dynamic pipeline scheduling to mask computation latency and maximize the utilization of heterogeneous computing resources.

\subsubsection{Pipeline Parallel Execution}
\label{sec:pipeline_parallel_exec}

Conventional synthetic data generation workflows predominantly rely on monolithic architectures where planning and rendering are tightly coupled within a synchronous execution loop. 
In this paradigm, the pipeline executes sequentially: Load Stage loads scenes, Plan Stage generates trajectories, Render Stage renders them, and then Store Stage executes persistence. 
This lockstep execution serializes hardware usage: GPUs idle during CPU-bound planning, and CPUs stall during GPU-bound rendering. 
Blocking storage I/O further amplifies these pipeline bubbles, preventing resource utilization.

\begin{figure}[t]
    \centering
    \includegraphics[width=\linewidth]{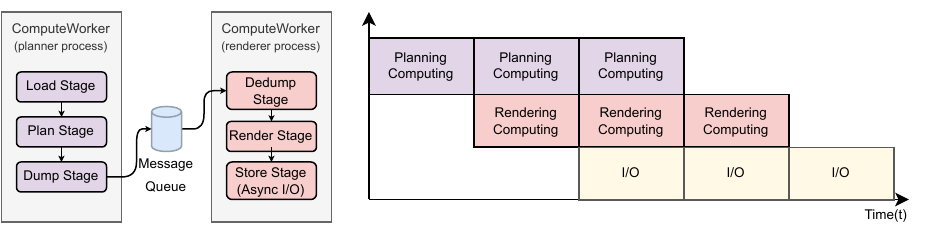}
    \caption{Decoupled Planning and Rendering Architecture. The decoupled architecture enables overlapping of Plan (CPU), Render (GPU), and Store (I/O) Stages.}
    \label{fig:decoupled_arch}
\end{figure}

To mitigate these bottlenecks, we propose a Pipeline Parallel Execution optimization that decouples the generation lifecycle into three asynchronous stages. 
As illustrated in Figure~\ref{fig:decoupled_arch}, we introduce intermediate buffers to sever the serial dependencies between stages.

\paragraph{Asynchronous Stage Decoupling.}
We bridge the decoupled stages using high-throughput message queues. 
The planner processes act as producers, pushing serialized simulation contexts into the queue. 
The renderer processes act as consumers, asynchronously fetching contexts for visualization. 
This design achieves pipeline parallelism: while the renderer process executes GPU-bound visualization for dequeued contexts, the planner process concurrently generates simulation states for subsequent tasks on the CPU. 
Moreover, this architecture supports the elastic deployment of multiple planner and renderer processes, allowing the system to align the aggregate processing rates of stages despite their inherent latency differences.

\paragraph{I/O Latency Hiding.}
To address the I/O bottleneck in the storage phase, we implement an Asynchronous Batch Storage mechanism. 
Rather than performing synchronous writes, the renderer process offloads data persistence to a dedicated I/O thread pool. 
By batching write operations, this design effectively isolates high-latency disk I/O from the critical rendering path.

\paragraph{Throughput Maximization.}
This pipelined architecture effectively masks stage-specific latencies. 
By overlapping CPU, GPU, and I/O operations, we transform the sparse, sequential execution timeline into a dense, parallel schedule. 
This ensures simultaneous saturation of heterogeneous hardware resources, including CPU cores, GPU compute units, and disk bandwidth, thereby significantly amplifying end-to-end generation throughput.

\subsubsection{Dynamic Pipeline Scheduling}
\label{sec:dynamic_scheduling}

Despite the benefits of pipeline parallelism, the latencies of the Plan and Render Stages remain inherently asymmetric. 
While static partitioning of planner and renderer processes can theoretically balance the pipeline, it faces significant practical limitations. 
First, it requires precise a priori profiling of stage latencies, which vary widely across tasks. 
Second, runtime stochasticity, such as planning failures where invalid trajectories are discarded, disrupts ideal computational overlap. 
Consequently, the Plan Stage often completes its workload prematurely, leaving resources idle while the Render Stage remains backlogged.

\begin{figure}
    \centering
    \includegraphics{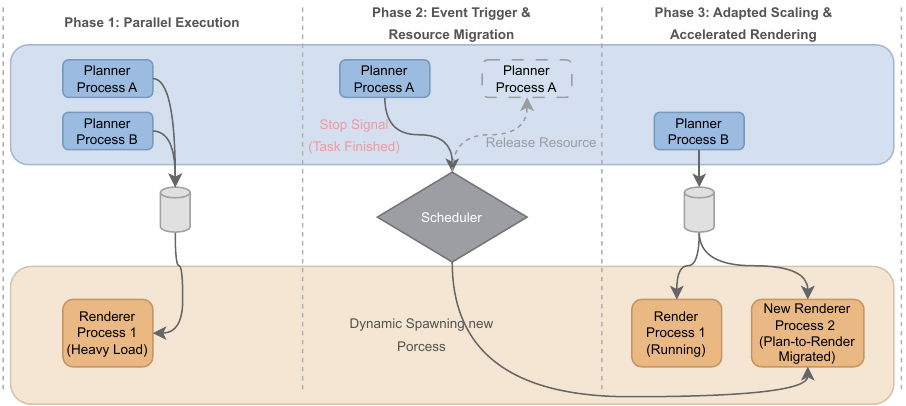}
    \caption{Dynamic Pipeline Scheduling Mechanism}
    \label{fig:dynamic_sched}
\end{figure}

To address these inefficiencies, we introduce a dynamic pipeline scheduling policy centered on adaptive resource reclamation. 
The mechanism is event-driven: when an upstream planner process exhausts its task stream, it transmits a termination signal, prompting the global Scheduler to immediately reclaim its resources. 
The Scheduler then evaluates the current queue backlog and dynamically provisions a new renderer process, which inherits the existing message queue connection and seamlessly joins the renderer group. 
Figure~\ref{fig:dynamic_sched} illustrates this process in a scenario initialized with two planner processes and one renderer process. 
As planner processes exit, their compute capacity is automatically reallocated to launch additional renderer processes, ensuring that resources flow fluidly from planning to rendering. 
This dynamic adjustment of stage parallelism effectively mitigates the long-tail latency observed in static configurations, yielding substantial improvements in end-to-end generation throughput.

\subsection{Schedule Opt Layer: Distributed Optimization}
\label{sec:distributed_opt}

To ensure cluster-scale efficiency and high availability, we implement a Distributed Optimization layer comprising the global load balancing and fault tolerance mechanisms.

\subsubsection{Global Load Balancing}
In distributed environments, static task assignment often induces severe load imbalance due to task heterogeneity and hardware performance variance (i.e., stragglers). 
To maximize cluster utilization, we implement a Global Load Balancer based on the Master-Worker architecture.

\begin{figure}
    \centering
    \includegraphics{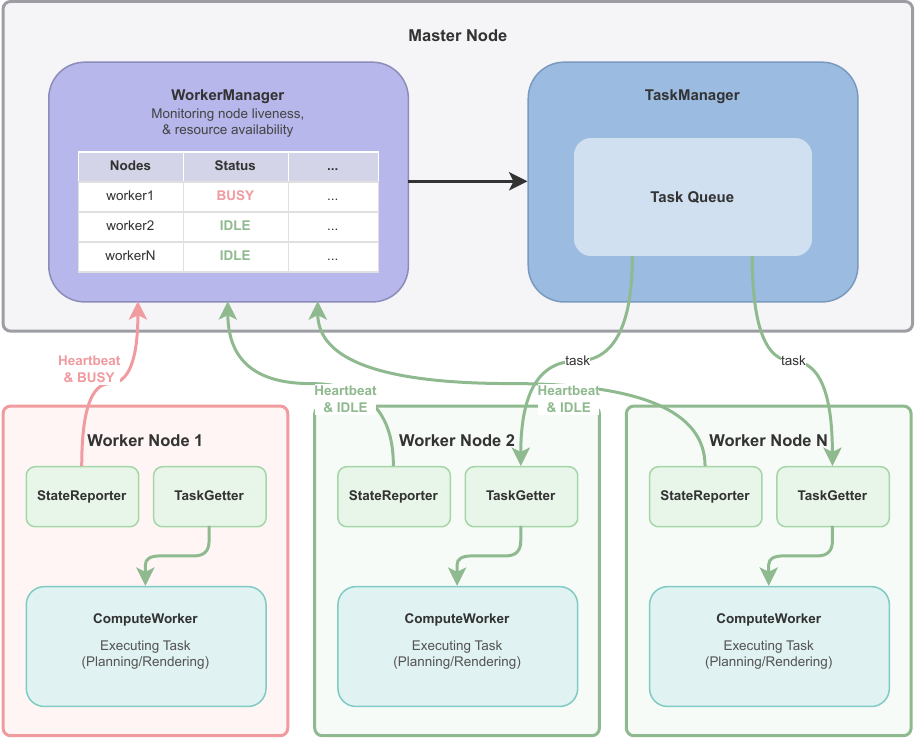}
    \caption{Architecture of the Global Load Balancer. The Master Node orchestrates task dispatching and state management, while Worker Nodes execute tasks and report status.}
    \label{fig:balance}
\end{figure}

As shown in Figure~\ref{fig:balance}, the Master Node serves as the central coordinator, maintaining global cluster state. 
It integrates a \texttt{WorkerManager} to track node liveness and resource availability, and a \texttt{TaskManager} to dynamically assign pending tasks to the most suitable workers based on real-time load metrics. 
The Worker Node hosts a \texttt{StateReporter} to push heartbeat updates to the Master and a \texttt{TaskGetter} to pull assignments. 
The \texttt{ComputeWorker} encapsulates the domain-specific execution logic (e.g., Planning or Rendering). 
Additionally, to mitigate scheduling overhead and network congestion, we employ a lazy context initialization strategy. 
Instead of transmitting full data payloads, the dispatcher sends only lightweight task metadata. 
Worker nodes lazily load the full execution context from shared storage only upon task initialization.

\subsubsection{Fault Tolerance via Supervisor}

We leverage Ray to manage the lifecycle of \texttt{ComputeWorker}s, providing basic availability via automatic process restarts. 
However, large-scale synthetic data generation remains susceptible to instability, particularly when incorporating complex physics simulators/renderers (e.g., Isaac Sim). 
These components often suffer from non-deterministic hangs or silent failures rather than immediate crashes, leading to execution stalls and resource leaks. 
Furthermore, conventional in-process monitoring is compromised by the Python Global Interpreter Lock (GIL), as deadlocks in the main execution thread frequently block monitoring threads.

To guarantee runtime robustness, we introduce an out-of-band \texttt{Supervisor}, implemented as an independent process decoupled from the \texttt{ComputeWorker}. 
Crucially, the \texttt{Supervisor} itself is managed by Ray, ensuring its own high availability.
This design establishes a robust failure detection loop.
Operationally, both the \texttt{ComputeWorker} and the \texttt{Supervisor} maintain their own \texttt{Status Monitor} components.
The \texttt{ComputeWorker} periodically updates its local \texttt{Status Monitor} status and synchronizes this state to the \texttt{Supervisor}'s \texttt{Status Monitor} via heartbeat messages.
The \texttt{Supervisor} continuously polls its own \texttt{Status Monitor} status, enforcing strict timeout policies to verify worker liveness.
Upon detecting a timeout, the \texttt{Supervisor} terminates the unresponsive \texttt{ComputeWorker} via \texttt{SIGKILL}. 
This triggers Ray to automatically respawn the worker and restore its execution context, effectively converting silent hangs into fail-stop errors and maintaining cluster stability without manual intervention.

\subsection{Backend Opt Layer: Renderer Optimization}
\label{sec:backend_opt}

To saturate the hardware capabilities of the underlying infrastructure, the Backend Opt Layer implements targeted renderer optimizations for the three primary renderers: Blender, Isaac Sim, and Gaussian Splatting.

\begin{figure}[t]
    \centering
    \includegraphics{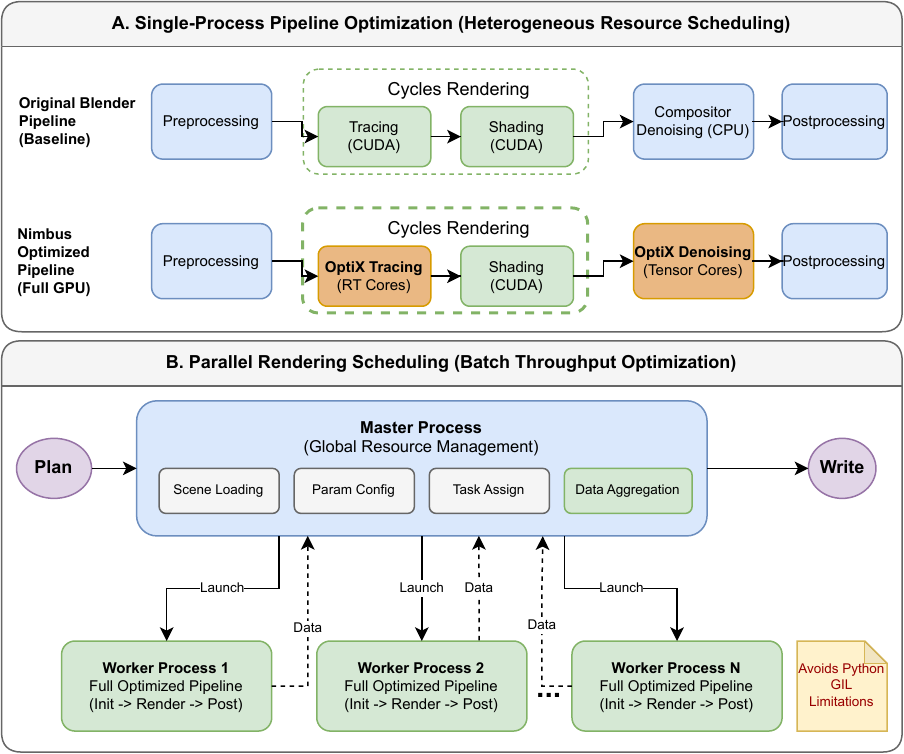}
    \caption{Blender Hardware-Accelerated Rendering Pipeline. The optimization delegates ray tracing and denoising to dedicated RT and Tensor Cores, respectively.}
    \label{fig:blender_opt}
\end{figure}

\subsubsection{Blender: Hardware-Accelerated Pipelining}
Blender is a ubiquitous open-source 3D creation suite for high-fidelity physically-based rendering.
We identify two critical bottlenecks in the InternData-N1 Blender baseline pipeline: the underutilization of specialized GPU hardware due to legacy execution paths, and the concurrency limits imposed by the GIL. 
We address these through heterogeneous compute offloading and multi-process parallelism.

\paragraph{Hardware-Aware Kernel Mapping.}
We re-engineer the rendering pipeline to exploit the specialized compute units of NVIDIA RTX architectures.
As depicted in Figure~\ref{fig:blender_opt}, the baseline pipeline suffers from resource contention on generic CUDA cores and incurs high latency from CPU-based denoising, which necessitates expensive device-to-host memory transfers.
To mitigate this, we implement a fully GPU-resident pipeline using the OptiX backend.
This design maps ray-triangle intersection kernels to dedicated RT Cores and delegates denoising to Tensor Cores via the OptiX AI Denoiser.
By confining the entire render-denoise loop to the GPU, we eliminate CPU bottlenecks and maximize the concurrent utilization of heterogeneous hardware resources.

\paragraph{Multi-Process Parallelism.}
To further maximize resource utilization on a single GPU, we implement a parallel rendering scheduling mechanism that launches multiple concurrent rendering processes in Nimbus Blender Renderer.
This design is motivated by the need to bypass the Python GIL, which limits standard Blender execution to a single thread and leaves GPU resources underutilized.
In this architecture, a master process acts as the global resource manager, responsible for scene loading, parameter configuration, task assignment, and final data aggregation.
It dispatches rendering tasks to a pool of independent worker processes, where each worker executes the full optimized pipeline (initialization, rendering, and post-processing) in isolation.
This process-level parallelism effectively circumvents the GIL, enabling the system to saturate the compute capacity of high-end GPUs.

\subsubsection{Isaac Sim: Stacked Rendering}
\begin{figure}[t]
    \centering
    \includegraphics[width=\linewidth]{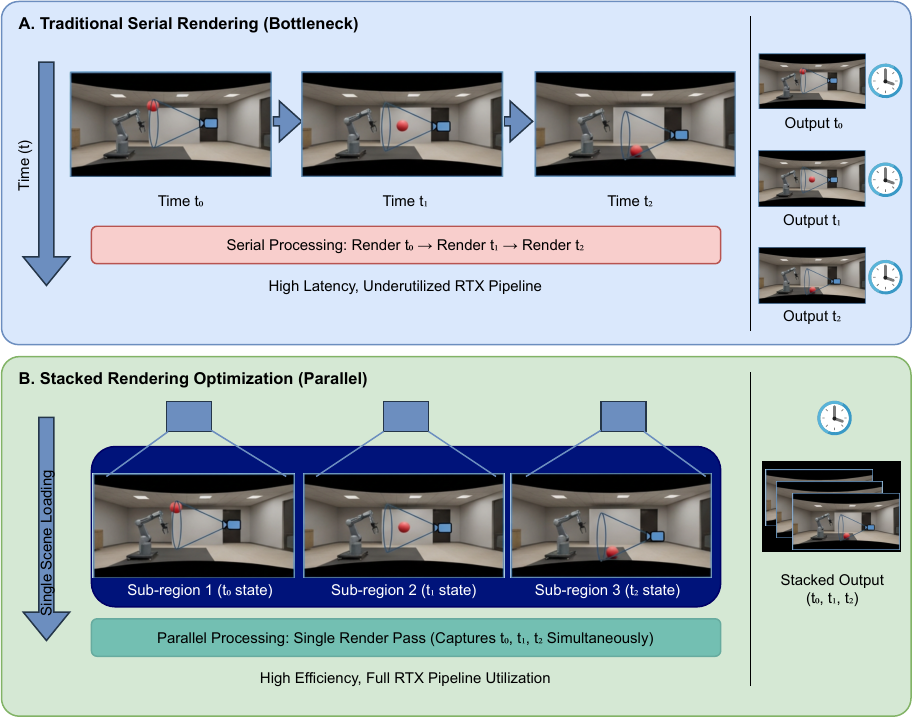}
    \caption{Isaac Sim Stacked Rendering Optimization. Temporal states are mapped to spatial sub-regions for single-pass rendering.}
    \label{fig:isaac_render}
\end{figure}

NVIDIA Isaac Sim is a high-fidelity robotics simulation and synthetic data generation tool built on the Omniverse platform. 
Designed specifically for embodied AI development, it serves as the core engine for physical simulation and multimodal data production within the synthetic data pipeline.
As illustrated in Figure~\ref{fig:isaac_render}, synthetic manipulation data generation typically requires recording temporal motion sequences of objects (e.g., the continuous process of a ball falling).
The conventional workflow employs serial state playback, where a single scene and camera render the object's different motion states sequentially over time ($t_0, t_1, t_2$).
This serial execution pattern fails to exploit the parallel capabilities of the RTX pipeline, creating a significant efficiency bottleneck.

To address this limitation, we introduce the \textit{Stacked Rendering} optimization, which maps multi-step temporal states (e.g., different phases of the falling ball) into multiple independent sub-regions within a single scene.
By deploying multiple cameras to capture these sub-regions in parallel, we enable multi-step data acquisition via a single scene load and one-pass rendering, effectively replacing the traditional serial temporal rendering approach.
This technique maximizes the throughput of the rendering pipeline.

\subsubsection{Gaussian Splatting: Kernel Fusion}
3D Gaussian Splatting is a novel 3D representation that models scenes as a collection of learnable 3D Gaussian primitives.
Each primitive encapsulates complete geometry (position, rotation, scale) and appearance (color, opacity) attributes.
The rendering pipeline projects these 3D Gaussians onto the image plane via splatting, executing a sequence of projection, tiling, sorting, and alpha blending to achieve real-time, high-quality scene synthesis.

To address the rendering efficiency bottlenecks in large-scale synthetic data generation, we integrate FlashGS~\cite{flashgs} and TC-GS~\cite{liao2025tc} to optimize the rasterization kernel.
FlashGS targets the computational and memory overheads of the standard pipeline.
It implements redundant Gaussian filtering to prune invalid or low-contribution primitives, parallelizes the rendering schedule, and enforces fine-grained GPU kernel execution control.
Furthermore, we leverage Tensor Cores to accelerate alpha blending, which dominates the computational cost in the rendering pipeline, as proposed in TC-GS.
These techniques significantly increase Gaussian Splatting rendering throughput, substantially improving the overall efficiency of the synthetic data generation pipeline.

\section{Evaluation}
\label{sec:experiment}

We evaluate Nimbus on an Alibaba Cloud cluster configured as detailed in Table~\ref{tab:hardware_config}.

\begin{table}[ht]
\centering
\caption{Hardware Configuration}
\label{tab:hardware_config}
\begin{tabular}{|l|l|}
\hline
\textbf{Component} & \textbf{Specification} \\
\hline
OS & Ubuntu 22.04 \\
GPU & NVIDIA RTX 4090 \\
CPU & Intel(R) Xeon(R) Gold 6462C @ 3889.285 MHz \\
\hline
\end{tabular}
\end{table}

We benchmark four data generation pipelines:

\begin{itemize}
\item \textbf{Nav-GS} (InternData-N1 Gaussian Splatting Pipeline): A navigation data collection pipeline that performs planning and rendering on coordinate-aligned mesh models and 3D Gaussian models. The scenes are custom indoor environments with approximately 3 million Gaussians per model.

\item \textbf{Nav-Mesh} (InternData-N1 Blender Pipeline): A traditional pipeline sampling endpoints from the 3D-FRONT dataset. It executes A-star path planning and utilizes Blender for rendering.

\item \textbf{GenManip} (InternData-M1 Pipeline): Executes object pick-and-place tasks on the Objaverse dataset.

\item \textbf{SimBox} (InternData-A1 Pipeline): Executes trash classification tasks on the Objaverse dataset.
\end{itemize}

\subsection{Performance Evaluation}

We evaluate the end-to-end performance of Nimbus against the unoptimized Baseline. 
Our evaluation spans four distinct pipelines: Nav-GS and Nav-Mesh for navigation, and GenManip and SimBox for manipulation.
We generate 150 trajectories for Nav-GS, 450 for Nav-Mesh (150 per scene), and 200 each for GenManip and SimBox. 
Given the inherent heterogeneity in task complexity and data volume, we focus on relative speedups (Baseline vs. Nimbus) within each pipeline rather than absolute cross-pipeline comparisons.

\subsubsection{End-to-End Throughput Analysis}

\begin{figure}[t]
    \centering
    \includegraphics{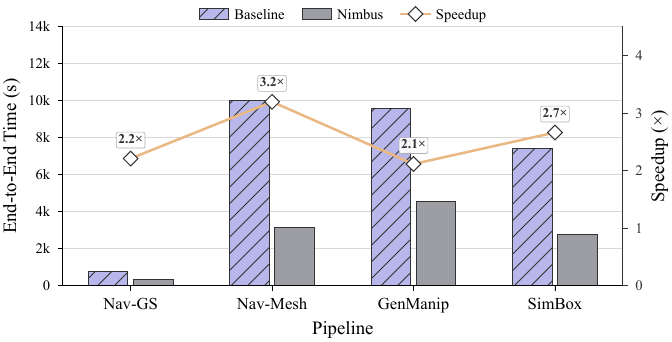}
    \caption{End-to-end performance comparison. Nimbus achieves $2.1\times$--$3.2\times$ speedup across all pipelines.}
    \label{fig:perf_comparison}
\end{figure}

\begin{figure}[t]
    \centering
    \includegraphics{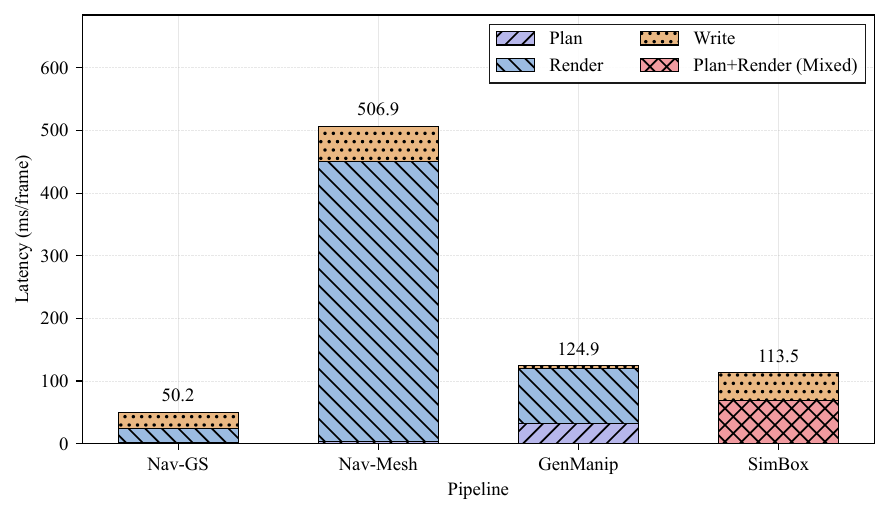}
    \caption{Latency breakdown of the Baseline pipeline. The monolithic design forces serial execution of Plan and Render stages.}
    \label{fig:baseline_latency}
\end{figure}

\begin{figure}[t]
    \centering
    \includegraphics{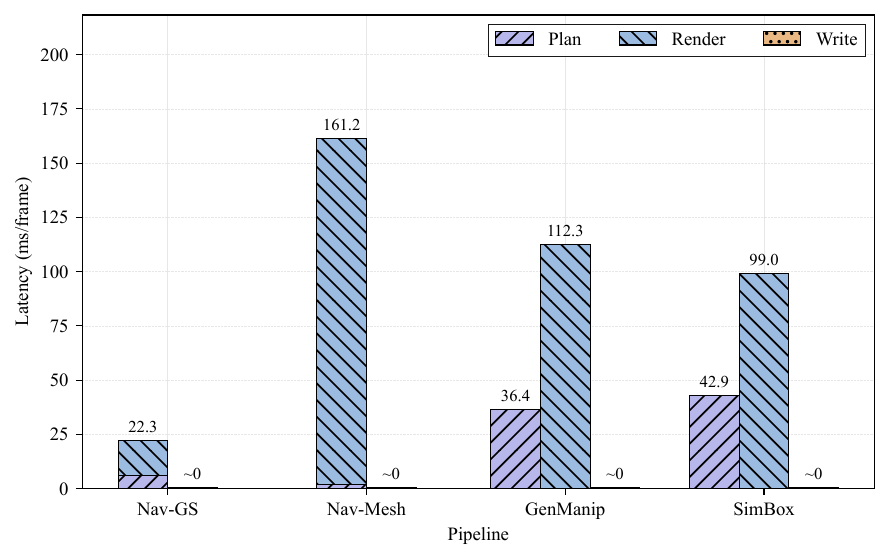}
    \caption{Latency breakdown of Nimbus. Under dynamic pipeline parallelism, end-to-end latency is bounded by the bottleneck stage rather than the cumulative sum. Values in the figure reflect single-worker latency and dynamic pipeline parallelism further scales effective throughput via intra-stage parallelism.}
    \label{fig:nimbus_latency}
\end{figure}

Figure~\ref{fig:perf_comparison} presents the end-to-end latency comparison. 
Nimbus delivers substantial performance gains across all workloads, achieving speedups ranging from $2.1\times$ to $3.2\times$.
To identify the sources of these gains, we decompose the latency profiles of key stages.
As shown in Figure~\ref{fig:baseline_latency}, the Baseline operates in a sequential mode where planning and rendering are tightly coupled.
In contrast, Nimbus (Figure~\ref{fig:nimbus_latency}) decouples these stages.
Crucially, under our pipeline parallelism and dynamic scheduling, the end-to-end latency is no longer the additive sum of individual stage latencies, but is instead dictated by the bottleneck stage.

In navigation tasks, the plan stage is computationally lightweight while the render stage and store stage dominate. 
Introducing a fully decoupled pipeline here would incur serialization and IPC overheads that negate potential concurrency benefits. 
Thus, Nimbus maintains synchronous execution for planning and rendering, focusing instead on two targeted optimizations:
\begin{itemize}
    \item \textbf{I/O Masking:} We decouple the storage stage. By offloading persistence to an asynchronous writer, we completely mask the $\sim$56 ms/frame disk I/O latency.
    \item \textbf{Renderer Optimization:} We optimize the critical rendering path via the Backend Opt Layer. For instance, Nav-Mesh rendering latency is reduced by $64\%$ (from 446.29 ms to 159.46 ms).
\end{itemize}

For GenManip and SimBox, Nimbus exploits the structural separation between CPU-bound planning and GPU-bound rendering:
\begin{itemize}
    \item \textbf{Pipeline Overlap:} By isolating planning and rendering into distinct \texttt{ComputeWorker}s, we overlap their execution. 
    The planning cost is effectively amortized within the rendering window.
    To further reduce the bottleneck latency, we increase rendering concurrency via multiple renderer \texttt{ComputeWorker}s.
    \item \textbf{Dynamic Resource Reallocation:} To mitigate tail latency, the scheduler dynamically reclaims resources from completed Planners to spawn additional Renderers. This adaptive reallocation ensures that computing resources remain saturated, minimizing idle time during the rendering tail.
\end{itemize}

\subsubsection{Theoretical Analysis of Effective Throughput}

While Nimbus achieves $2\times$--$3\times$ speedups, stage-wise latencies (Figure~\ref{fig:nimbus_latency}) show that total per-frame computation (Plan and Render) remains comparable to the Baseline. This indicates that performance gains stem from architectural efficiency rather than operation reduction. We formalize this via Effective Stage Throughput.

\paragraph{Throughput Model}
In a decoupled architecture, stage performance is defined by per-frame latency $\ell_s$ and time-averaged concurrency $\bar{N}_s = \frac{1}{T}\int_0^T N_s(t)dt$. The effective throughput is:
$$
\mu_s = \frac{\bar{N}_s}{\ell_s} \quad (\text{frames/s}).
$$
The system's steady-state theoretical maximum throughput is $\lambda_{\text{theory}} = \min(\mu_{\text{plan}}, \mu_{\text{render}})$. Unlike the Baseline's degenerate pipeline ($\lambda_{\text{base}} \approx (\sum \ell_i)^{-1}$), Nimbus maximizes $\lambda_{\text{theory}}$ by increasing $\bar{N}_s$ and hiding latencies via overlap.

\paragraph{Impact of dynamic pipeline scheduling}
In manipulation tasks where $\ell_{\text{render}} \gg \ell_{\text{plan}}$, the system is GPU-bound. Nimbus initializes with a balanced Planner:Renderer ratio to saturate queues, then dynamically reallocates resources. As Planners exit, $\bar{N}_{\text{plan}}$ decreases while $\bar{N}_{\text{render}}$ increases, boosting $\mu_{\text{render}}$ exactly when the bottleneck shifts to the tail.

\paragraph{Correlation Between Failure Rates and Latency}
A counter-intuitive dynamic exists in GPU-bound regimes: higher planning success rates correlate with longer total execution times. Let $X_i \in \{0, 1\}$ be the success indicator for attempt $i$. The total execution time is approximated by:
$$
T_{\text{total}}^{\text{Nimbus}} \approx \frac{\ell_{\text{render}}^{\text{Nimbus}}}{\bar{N}_{\text{render}}} \sum_{i=1}^{M} X_i.
$$
Since failed plans are pruned before rendering, a higher failure rate reduces the total rendering load. To evaluate efficiency fairly, we define \textbf{Effective Successful Throughput} $\lambda_{\text{succ}} = (\sum X_i) / T_{\text{total}}$. Nimbus maintains high $\lambda_{\text{succ}}$ robustly across varying failure rates because dynamic pipeline scheduling ensures that the throughput of successful samples approaches $\lambda_{\text{theory}}$.

\subsection{Scalability}

When scaling tasks to larger magnitudes, additional compute nodes are required. This inevitably introduces two critical challenges: global load balancing and fault recovery for complex physics simulators. Both challenges significantly impact the framework's scalability.

We address load imbalance arising from task heterogeneity and hardware performance variations through a master-worker architecture with a global load balancer, and ensure runtime robustness via a Supervisor mechanism, thereby achieving excellent scalability.

\begin{figure}[ht]
    \centering
    \includegraphics[width=\linewidth]{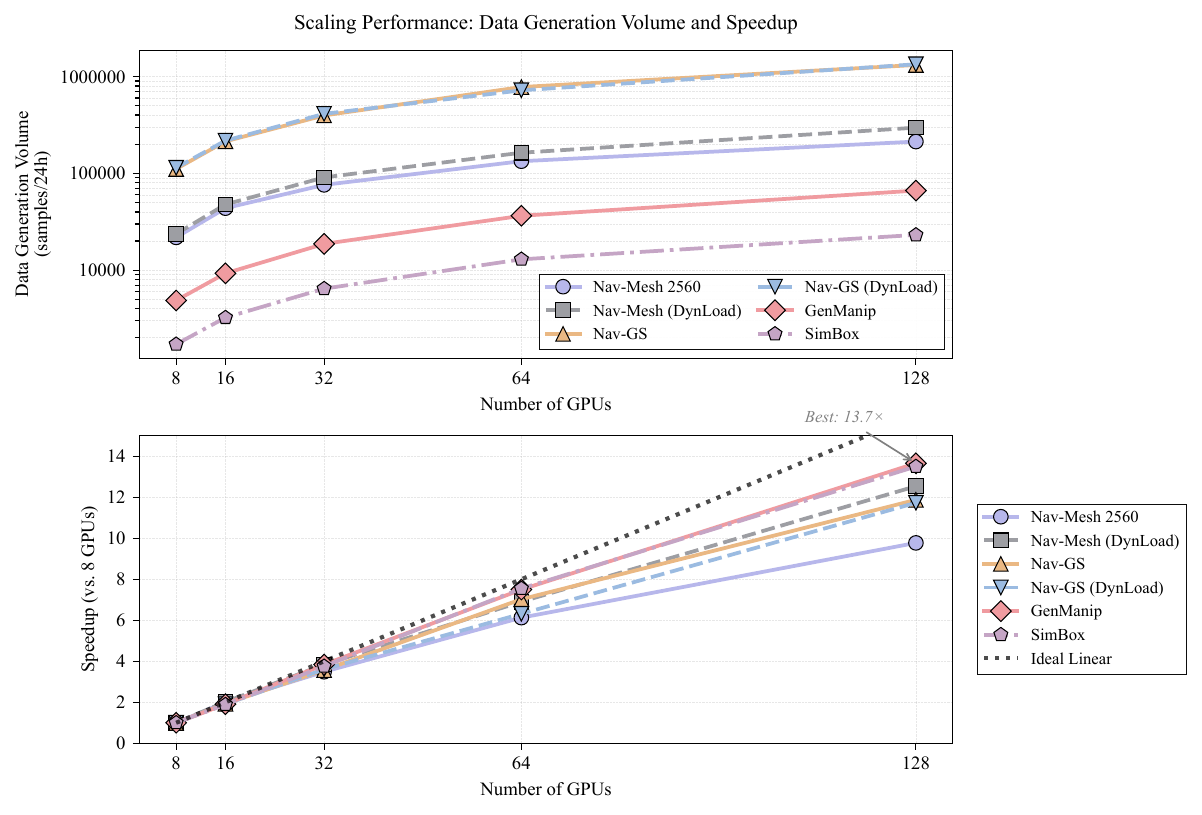}
    \caption{24-Hour Data Generation Volume and Scaling Factors}
    \label{fig:scaling_results}
\end{figure}

We evaluate the system's scalability by conducting 24-hour data generation tasks across clusters ranging from 8 to 128 GPUs. Notably, for the Nav-Mesh pipeline, our evaluation comprises 2,560 distinct scenes. We also conduct comparative experiments with the dynamic load balancing parameter (\texttt{dynload}) enabled and disabled, where enabling \texttt{dynload} activates the global load balancer. The results are shown in Figure~\ref{fig:scaling_results}.

Theoretically, throughput should scale linearly with the number of GPUs. Our experimental results demonstrate that when scaling from 8 to 128 GPUs, the system achieves approximately 86\% linear scaling efficiency, indicating excellent scalability. Throughout the entire 24-hour testing period, the system maintained stable operation with no node failures or task failures, demonstrating exceptional robustness.

The deviation from perfect linear scaling can be attributed to two primary factors. First, task execution exhibits inherent tail effects. Even with load balancing, the completion time of the final batch of tasks affects overall throughput measurements, and this effect becomes more pronounced as the number of nodes increases. Second, the scheduling efficiency of the load balancer is influenced by task granularity. As the number of nodes increases, the number of tasks assigned to each node decreases correspondingly, which reduces the scheduling flexibility available to the global load balancer. In our 128-GPU configuration, the 2,560 scenes are distributed with only approximately 20 scenes per GPU on average. This relatively small task pool per node is insufficient for the load balancer to fully exploit its dynamic scheduling capabilities and effectively mitigate the inevitable tail effects, thereby limiting overall resource utilization efficiency.

Notably, despite these factors, the system maintains high resource utilization on large-scale clusters, benefiting from the effective coordination between our master-worker architecture and the Supervisor mechanism. This result validates that our architectural design can support large-scale data generation requirements in production environments.
\section{Conclusion}
\label{sec:conclusion}

Data scarcity remains the primary bottleneck in training embodied intelligence models. While synthetic data offers a viable path forward, existing generation pipelines are plagued by fragmentation, inefficiency, and instability. In this paper, we presented Nimbus, a unified framework that resolves these architectural deficiencies through a systematic design. By consolidating navigation and manipulation workflows into a single framework, Nimbus successfully orchestrated the large-scale construction of the InternData dataset series. Our evaluation demonstrates that the system achieves a 2--3$\times$ throughput improvement over baselines via dynamic pipeline scheduling and rendering optimizations. Furthermore, its fault-tolerant design ensures robust, continuous operation on large-scale GPU clusters, satisfying the stability requirements for massive data campaigns.

By replacing ad hoc scripts with a standardized, high-performance infrastructure, Nimbus circumvents the prohibitive costs of physical data collection. It delivers the high-capacity, multimodal data streams necessary to advance manipulation and navigation models. Looking ahead, we plan to extend the framework in three directions: integrating diverse trajectory planners to enrich behavioral variance, incorporating automated scene generation to broaden environmental compatibility, and implementing automated task configuration to further enhance the generalization capabilities of downstream models.

\bibliographystyle{unsrt}
\bibliography{main}

@String{Computing = "Computing" }

@String{Computer = "{IEEE} Computer" }

@String{Chelsea = "Chelsea" }

@article{objectnav,
  title={Object goal navigation using goal-oriented semantic exploration},
  author={Chaplot, Devendra Singh and Gandhi, Dhiraj Prakashchand and Gupta, Abhinav and Salakhutdinov, Russ R},
  journal={Advances in Neural Information Processing Systems},
  volume={33},
  pages={4247--4258},
  year={2020}
}

@inproceedings{chang2017matterport3d,
  title={Matterport3D: Learning from RGB-D Data in Indoor Environments},
  author={Chang, Angel and Dai, Angela and Funkhouser, Thomas and Halber, Maciej and Niessner, Matthias and Savva, Manolis and Song, Shuran and Zeng, Andy and Zhang, Yinda},
  booktitle={International Conference on 3D Vision (3DV)},
  year={2017}
}

@inproceedings{hm3d,
  title={Habitat-Matterport 3D Dataset (HM3D): 1000 Large-scale 3D Environments for Embodied AI},
  author={Ramakrishnan, Santhosh Kumar and Gokaslan, Aaron and Wijmans, Erik and Maksymets, Oleksandr and Clegg, Alexander and Turner, John and Undersander, Eric and Galuba, Wojciech and Westbury, Andrew and Chang, Angel and Savva, Manolis and Zhao, Yili and Batra, Dhruv},
  booktitle={Advances in Neural Information Processing Systems (NeurIPS)},
  year={2021}
}

@inproceedings{anderson2018vision,
  title={Vision-and-Language Navigation: Interpreting Visually-Grounded Navigation Instructions in Real Environments},
  author={Anderson, Peter and Wu, Qi and Teney, Damien and Bruce, Jake and Johnson, Mark and S{\"u}nderhauf, Niko and Reid, Ian and Gould, Stephen and van den Hengel, Anton},
  booktitle={IEEE Conference on Computer Vision and Pattern Recognition (CVPR)},
  year={2018}
}

@inproceedings{ku2020room,
  title={Room-Across-Room: Multilingual Vision-and-Language Navigation with Dense Spatiotemporal Grounding},
  author={Ku, Alexander and Anderson, Peter and Patel, Roma and Ie, Eugene and Baldridge, Jason},
  booktitle={Empirical Methods in Natural Language Processing (EMNLP)},
  year={2020}
}

@inproceedings{thomason2019vision,
  title={Vision-and-Dialog Navigation},
  author={Thomason, Jesse and Murray, Michael and Chakravarty, Maya and Zettlemoyer, Luke},
  booktitle={Conference on Robot Learning (CoRL)},
  year={2019}
}

@inproceedings{qi2020reverie,
  title={REVERIE: Remote Embodied Visual Referring Expression in Real Indoor Environments},
  author={Qi, Yuankai and Wu, Qi and Anderson, Peter and Wang, Xin and Wang, William Yang and Shen, Chunhua and van den Hengel, Anton},
  booktitle={IEEE Conference on Computer Vision and Pattern Recognition (CVPR)},
  year={2020}
}

@inproceedings{padalkar2023open,
  title={Open X-Embodiment: Robotic Learning Datasets and RT-X Models},
  author={Padalkar, Abhishek and Pooley, Acorn and Jain, Ajinkya and Bewley, Alex and Herzog, Alex and Irpan, Alex and Khazatsky, Alexander and Rai, Anikait and Singh, Anikait and Brohan, Anthony and others},
  booktitle={IEEE International Conference on Robotics and Automation (ICRA)},
  year={2024}
}

@article{black2024pi0,
  title={$\pi_0$: A Vision-Language-Action Flow Model for General Robot Control},
  author={Black, Kevin and Brown, Noah and Driess, Danny and Esmail, Adnan and Equi, Michael and Finn, Chelsea and Fusai, Niccolo and Groom, Lachy and Hausman, Karol and Ichter, Brian and others},
  journal={arXiv preprint arXiv:2410.24164},
  year={2024}
}

@article{wu2024robomind,
  title={RoboMIND: Benchmark on Multi-embodiment Intelligence Normative Data for Robot Manipulation},
  author={Wu, Kun and Hou, Chengkai and Liu, Jiaming and Che, Zhengping and Ju, Xiaozhu and Yang, Zhuqin and Li, Meng and Zhao, Yinuo and Xu, Zhiyuan and Yang, Guang and others},
  journal={arXiv preprint arXiv:2412.13877},
  year={2024}
}

@article{bu2025agibot,
  title={AgiBot World Colosseo: A Large-scale Manipulation Platform for Scalable and Intelligent Embodied Systems},
  author={Bu, Qingwen and Chen, Yilun and others},
  journal={arXiv preprint arXiv:2503.06669},
  year={2025}
}

@article{jiang2025galaxea,
  title={Galaxea Open-World Dataset and G0 Dual-System VLA Model},
  author={Jiang, Tao and Yuan, Tianyuan and Liu, Yicheng and Lu, Chenhao and Cui, Jianning and Liu, Xiao and Cheng, Shuiqi and Gao, Jiyang and Xu, Huazhe and Zhao, Hang},
  journal={arXiv preprint arXiv:2509.00576},
  year={2025}
}

@inproceedings{jiang2025iris,
  title={IRIS: An Immersive Robot Interaction System},
  author={Jiang, Xinkai and Yuan, Qihao and Dincer, Enes Ulas and Zhou, Hongyi and Li, Ge and Li, Xueyin and Jia, Xiaogang and Schnizer, Timo and Schreiber, Nicolas and Liao, Weiran and others},
  booktitle={Conference on Robot Learning (CoRL)},
  year={2025}
}

@inproceedings{chi2024universal,
  title={Universal Manipulation Interface: In-The-Wild Robot Teaching Without In-The-Wild Robots},
  author={Chi, Cheng and Feng, Siyuan and Du, Yilun and Xu, Zhenjia and Cousineau, Eric and Burchfiel, Benjamin and Song, Shuran},
  booktitle={IEEE Conference on Computer Vision and Pattern Recognition (CVPR)},
  year={2024}
}

@inproceedings{zhaxizhuoma2025fastumi,
  title={FastUMI: A Scalable and Hardware-Independent Universal Manipulation Interface with Dataset},
  author={Zhaxizhuoma, Zhaxizhuom and Liu, Kehui and Guan, Chuyue and Jia, Zhongjie and Wu, Ziniu and Liu, Xin and Wang, Tianyu and Liang, Shuai and Chen, Pengan and others},
  booktitle={Conference on Robot Learning (CoRL)},
  year={2025}
}

@inproceedings{mandlekar2023mimicgen,
  title={MimicGen: A Data Generation System for Scalable Robot Learning using Human Demonstrations},
  author={Mandlekar, Ajay and Nasiriany, Soroush and Wen, Bowen and Akinola, Iretiayo and Narang, Yashraj and Fan, Linxi and Zhu, Yuke and Fox, Dieter},
  booktitle={Conference on Robot Learning (CoRL)},
  year={2023}
}

@inproceedings{nasiriany2024robocasa,
  title={RoboCasa: Large-Scale Simulation of Everyday Tasks for Generalist Robots},
  author={Nasiriany, Soroush and Maddukuri, Abhiram and Zhang, Lance and Parikh, Adeet and Lo, Aaron and Joshi, Abhishek and Mandlekar, Ajay and Zhu, Yuke},
  booktitle={Robotics: Science and Systems (RSS)},
  year={2024}
}

@inproceedings{gao2025genmanip,
  title={GenManip: LLM-driven Simulation for Generalizable Instruction-Following Manipulation},
  author={Gao, Ning and Chen, Yilun and Pang, Jiangmiao and others},
  booktitle={IEEE Conference on Computer Vision and Pattern Recognition (CVPR)},
  year={2025}
}

@article{james2020rlbench,
  title={RLBench: The Robot Learning Benchmark \& Learning Environment},
  author={James, Stephen and Ma, Zicong and Arrojo, David Rovick and Davison, Andrew J.},
  journal={IEEE Robotics and Automation Letters (RA-L)},
  year={2020}
}

@inproceedings{gu2023maniskill2,
  title={ManiSkill2: A Unified Benchmark for Generalizable Manipulation Skills},
  author={Gu, Jiayuan and Xiang, Fanbo and Li, Xuanlin and Ling, Zhan and Liu, Xiqiang and Sang, Tongzhou and Seymour, Ackermann and Wei, Xuan and Su, Hao},
  booktitle={International Conference on Learning Representations (ICLR)},
  year={2023}
}

@misc{airflow,
  title={Apache Airflow},
  author={{The Apache Software Foundation}},
  year={2024},
  howpublished={\url{https://airflow.apache.org/}}
}

@misc{prefect,
  title={Prefect: The Workflow Orchestration Platform},
  author={{Prefect Technologies, Inc.}},
  year={2024},
  howpublished={\url{https://www.prefect.io/}}
}

@article{generalist2025gen0,
author = {Generalist AI Team},
title = {GEN-0: Embodied Foundation Models That Scale with Physical Interaction},
journal = {Generalist AI Blog},
year = {2025},
note = {https://generalistai.com/blog/preview-uqlxvb-bb.html},
}

@misc{pi05,
      title={$\pi_{0.5}$: a Vision-Language-Action Model with Open-World Generalization}, 
      author={Physical Intelligence and Kevin Black and Noah Brown and James Darpinian and Karan Dhabalia and Danny Driess and Adnan Esmail and Michael Equi and Chelsea Finn and Niccolo Fusai and Manuel Y. Galliker and Dibya Ghosh and Lachy Groom and Karol Hausman and Brian Ichter and Szymon Jakubczak and Tim Jones and Liyiming Ke and Devin LeBlanc and Sergey Levine and Adrian Li-Bell and Mohith Mothukuri and Suraj Nair and Karl Pertsch and Allen Z. Ren and Lucy Xiaoyang Shi and Laura Smith and Jost Tobias Springenberg and Kyle Stachowicz and James Tanner and Quan Vuong and Homer Walke and Anna Walling and Haohuan Wang and Lili Yu and Ury Zhilinsky},
      year={2025},
      eprint={2504.16054},
      archivePrefix={arXiv},
      primaryClass={cs.LG},
      url={https://arxiv.org/abs/2504.16054}, 
}

@misc{interndata2025a1,
      title={InternData-A1: Pioneering High-Fidelity Synthetic Data for Pre-training Generalist Policy}, 
      author={Yang Tian and Yuyin Yang and Yiman Xie and Zetao Cai and Xu Shi and Ning Gao and Hangxu Liu and Xuekun Jiang and Zherui Qiu and Feng Yuan and Yaping Li and Ping Wang and Junhao Cai and Jia Zeng and Hao Dong and Jiangmiao Pang},
      year={2025},
      eprint={2511.16651},
      archivePrefix={arXiv},
      primaryClass={cs.RO},
      url={https://arxiv.org/abs/2511.16651}, 
}

@misc{internvla-n1,
    title = {{InternVLA-N1: An} Open Dual-System Navigation Foundation Model with Learned Latent Plans},
    author = {InternNav Team},
    year = {2025},
    booktitle={arXiv},
}

@misc{streamvln,
    title = {StreamVLN: Streaming Vision-and-Language Navigation via SlowFast Context Modeling},
    author = {Wei, Meng and Wan, Chenyang and Yu, Xiqian and Wang, Tai and Yang, Yuqiang and Mao, Xiaohan and Zhu, Chenming and Cai, Wenzhe and Wang, Hanqing and Chen, Yilun and Liu, Xihui and Pang, Jiangmiao},
    booktitle={arXiv},
    year = {2025}
}

@misc{navdp,
      title={NavDP: Learning Sim-to-Real Navigation Diffusion Policy with Privileged Information Guidance}, 
      author={Wenzhe Cai and Jiaqi Peng and Yuqiang Yang and Yujian Zhang and Meng Wei and Hanqing Wang and Yilun Chen and Tai Wang and Jiangmiao Pang},
      year={2025},
      eprint={2505.08712},
      archivePrefix={arXiv},
      primaryClass={cs.RO},
      url={https://arxiv.org/abs/2505.08712}, 
}

@misc{internvla-m1,
      title={InternVLA-M1: A Spatially Guided Vision-Language-Action Framework for Generalist Robot Policy}, 
      author={InternVLA-M1 contributors},
      year={2025},
      eprint={2510.13778},
      archivePrefix={arXiv},
      primaryClass={cs.RO},
      url={https://arxiv.org/abs/2510.13778}, 
}

@inproceedings{
qi2024zero,
title={Zero Bubble (Almost) Pipeline Parallelism},
author={Penghui Qi and Xinyi Wan and Guangxing Huang and Min Lin},
booktitle={The Twelfth International Conference on Learning Representations},
year={2024},
url={https://openreview.net/forum?id=tuzTN0eIO5}
}

@inproceedings{megatron,
author = {Narayanan, Deepak and Shoeybi, Mohammad and Casper, Jared and LeGresley, Patrick and Patwary, Mostofa and Korthikanti, Vijay and Vainbrand, Dmitri and Kashinkunti, Prethvi and Bernauer, Julie and Catanzaro, Bryan and Phanishayee, Amar and Zaharia, Matei},
title = {Efficient large-scale language model training on GPU clusters using megatron-LM},
year = {2021},
isbn = {9781450384421},
publisher = {Association for Computing Machinery},
address = {New York, NY, USA},
url = {https://doi.org/10.1145/3458817.3476209},
doi = {10.1145/3458817.3476209},
booktitle = {Proceedings of the International Conference for High Performance Computing, Networking, Storage and Analysis},
articleno = {58},
numpages = {15},
location = {St. Louis, Missouri},
series = {SC '21}
}

@inproceedings{wang2024dexcap,
  title={DexCap: Scalable and Portable Mocap Data Collection System for Dexterous Manipulation},
  author={Wang, Chen and Shi, Haochen and Wang, Weizhuo and Zhang, Ruohan and Fei-Fei, Li and Liu, Karen},
  booktitle={RSS 2024 Workshop: Data Generation for Robotics},
  year = {2024}
}

@inproceedings{xudexumi,
  title={DexUMI: Using Human Hand as the Universal Manipulation Interface for Dexterous Manipulation},
  author={Xu, Mengda and Zhang, Han and Hou, Yifan and Xu, Zhenjia and Fan, Linxi and Veloso, Manuela and Song, Shuran},
  booktitle={3rd RSS Workshop on Dexterous Manipulation: Learning and Control with Diverse Data},
  year = {2025}
}

@InProceedings{flashgs,
    author    = {Feng, Guofeng and Chen, Siyan and Fu, Rong and Liao, Zimu and Wang, Yi and Liu, Tao and Hu, Boni and Xu, Linning and Pei, Zhilin and Li, Hengjie and Li, Xiuhong and Sun, Ninghui and Zhang, Xingcheng and Dai, Bo},
    title     = {FlashGS: Efficient 3D Gaussian Splatting for Large-scale and High-resolution Rendering},
    booktitle = {Proceedings of the IEEE/CVF Conference on Computer Vision and Pattern Recognition (CVPR)},
    month     = {June},
    year      = {2025},
    pages     = {26652-26662}
}

@inproceedings{moritz2018ray,
  title={Ray: A distributed framework for emerging $\{$AI$\}$ applications},
  author={Moritz, Philipp and Nishihara, Robert and Wang, Stephanie and Tumanov, Alexey and Liaw, Richard and Liang, Eric and Elibol, Melih and Yang, Zongheng and Paul, William and Jordan, Michael I and others},
  booktitle={13th USENIX symposium on operating systems design and implementation (OSDI 18)},
  pages={561--577},
  year={2018}
}

@Article{kerbl3Dgaussians,
      author       = {Kerbl, Bernhard and Kopanas, Georgios and Leimk{\"u}hler, Thomas and Drettakis, George},
      title        = {3D Gaussian Splatting for Real-Time Radiance Field Rendering},
      journal      = {ACM Transactions on Graphics},
      number       = {4},
      volume       = {42},
      month        = {July},
      year         = {2023},
      url          = {https://repo-sam.inria.fr/fungraph/3d-gaussian-splatting/}
}

@article{liao2025tc,
  title={Tc-gs: A faster gaussian splatting module utilizing tensor cores},
  author={Liao, Zimu and Ding, Jifeng and Cui, Siwei and Gong, Ruixuan and Hu, Boni and Wang, Yi and Li, Hengjie and Zhang, XIngcheng and Wang, Hui and Fu, Rong},
  journal={to appear in the SIGGRAPH Asia 2025 Conference Proceedings},
  year={2025}
}

@misc{internvla-a1,
      title={InternVLA-A1: Unifying Understanding, Generation and Action for Robotic Manipulation}, 
      author={InternVLA-A1 contributors},
      year={2026},
      eprint={2601.02456},
      archivePrefix={arXiv},
      primaryClass={cs.RO},
      url={https://arxiv.org/abs/2601.02456}, 
}

\end{document}